\documentclass[letterpaper]{article} 
\usepackage{aaai23}  
\usepackage{times}  
\usepackage{helvet}  
\usepackage{courier}  
\usepackage[hyphens]{url}  
\usepackage{graphicx} 
\usepackage{natbib}  
\usepackage{caption} 
\DeclareCaptionStyle{ruled}{labelfont=normalfont,labelsep=colon,strut=off} 
\usepackage{algorithm}
\usepackage{algorithmic}
\usepackage{amsmath}
\usepackage{amsfonts}
\usepackage{multirow}
\usepackage{booktabs}
\usepackage{amssymb}

\usepackage{newfloat}
\usepackage{listings}
\floatstyle{ruled}
\newfloat{listing}{tb}{lst}{}
\floatname{listing}{Listing}
\nocopyright
\title{Refined Vision-Language Modeling for Fine-grained Multi-modal Pre-training}
\author{
    Lisai Zhang$^1$,
    Qingcai Chen$^1$,
    Zhijian Chen$^2$,
    Zhonghua Li$^2$,
    Yunpeng Han$^1$,
    Zhao Cao$^2$
}
\affiliations{
    $^1$Harbin Institute of Technology, Shenzhen, \ \
    $^2$Huawei Technology. Inc 
}

\usepackage{bibentry}

\begin{document}

\maketitle

\begin{abstract}
Fine-grained supervision based on object annotations has been widely used for vision and language pre-training (VLP). However, in real-world application scenarios, aligned multi-modal data is usually in the image-caption format, which only provides coarse-grained supervision. It is not only cost-expensive but also compute-expensive to collect object annotations and build object annotation pre-extractor for different scenarios. In this paper, we propose a fine-grained VLP scheme without object annotations from the linguistic perspective. First, we propose a homonym sentence rewriting (HSR) algorithm to provide token-level supervision. The algorithm replaces a verb/noun/adjective/quantifier word of the caption with its homonyms from WordNet. Correspondingly, we propose refined vision-language modeling (RVLM) framework to exploit the token-level supervision. Three refined tasks, i.e., refined image-text contrastive (RITC), refined image-text matching (RITM), and replace language modeling (RLM) are proposed to learn the fine-grained alignment.   Extensive experiments on several downstream tasks demonstrate the superior performance of the proposed method.
\end{abstract}

\section{Introduction}
Vision and language pre-training (VLP) greatly empower machines' ability to understand multi-modal data in recent years. VLP aims to understand multi-modal semantics like a human, who is able to recognize fine-grained alignment between image objects and tokens. However, in real-world application scenarios, aligned multi-modal data is usually in the coarse-grained image-caption format, it is an important challenge to learn fine-grained alignment from the coarse-grained supervision.
 
\begin{figure}
	\includegraphics[width=\linewidth]{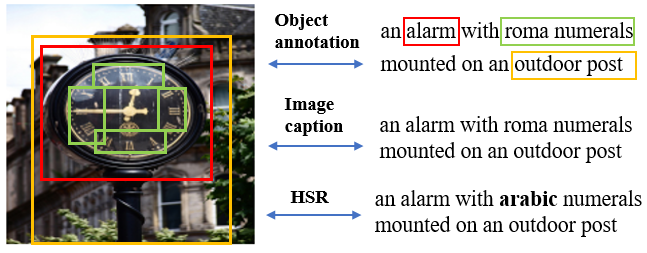}
	\caption{Illustration of the object annotation supervision, image caption supervision, and our proposed homonyms sentence rewriting (HSR) supervision schemes. Combined with the ground truth text, HSR provides token-level supervision without human labeling.}
	\label{fig.idea}
\end{figure}

Most of the existing fine-grained VLP methods~\cite{XITM,MVPTR} learns cross-modal semantic alignment from an object-tag or object-phrase style supervision as Fig.~\ref{fig.idea} (a). Since such fine-grained annotations are not available for the pre-training corpus, a common paradigm is extracting the object region features and their corresponding tags~\cite{vlbert,imagebert,oscar,vinvl} through off-the-shelf object detection models like Faster-RCNN~\cite{fasterRCNN:2015}. 
Afterward, the object features are fed to the Transformer~\cite{transformer} as a visual token, and self-supervision tasks such as masked language modeling (MLM) and image-text matching (ITM) are used to learn the multimodal representation.  
Recent method~\cite{MVPTR} extends the object-tag annotation to multi-grained objects-concepts using scene graph parser~\cite{scene} and achieves significant performance improvement. 
A recent study has proved~\cite{vinvl} that the effectiveness of these VLP methods highly benefits from the object detector quality. 
However, these off-the-shelf pre-extractors are built for certain scenes upon expensive bounding box annotations, which inevitably limits VLP's application to more general domains.
On the other hand, the object detector and scene graph parser need to infer on high-resolution images, which is proved~\cite{ALBEF} to be compute-expensive.

A weakly-supervised  paradigm~\cite{fashionbert,vilt} to achieve fine-grained VLP without object annotations is regarding the image patches as visual tokens. However, removal of the object detector is reported~\cite{vilt} to achieve lower performance. 
Recent work~\cite{ALBEF,li2022blip,coca} use image-text contrastive (ITC) objective to the uni-modal encoder before fusing the multi-modal features. Together with MLM and ITM objectives on the multi-modal encoder, these models achieve excellent performance.  However, it has been reported~\cite{MVPTR} that these methods still fall short of the fine-grained approach when using the same order of magnitude training corpus.

Semantics of the vision modality, e.g.,textures, objects, and actions, are not naturally segmented from others, therefore it is hard to develop explicit fine-grained supervision from the vision modality. 
Fortunately, language symbols are naturally discrete, whose basic linguistic semantic units are tokens. It would be possible to create fine-grained self-supervision signals from language modality without human labeling. 
The MLM is a successful supervision from this perspective.
However, most VLP methods are simply using the same MLM design as the BERT~\cite{bert}.
We argue that the language modeling supervision in VLP still lacks systematical exploration.

In this paper, we propose a refined vision-language modeling (RVLM) scheme for VLP. First, we propose a homonym sentence rewriting algorithm (HSR) to produce fine-grained supervision. As illustrated in Fig.~\ref{fig.idea}, we replace one verb/noun/adjective/quantifier token of a caption by it's homonyms from the WordNet~\cite{wordnet}. The rewritten sentence is used as a negative sample, who's semantic differs from the original caption by only one token. 
Afterwards, we design three self-supervised objectives to capture the supervision. On the uni-modal encoders, a refined image-text contrastive (RITC) task maximizes the similarity between the image and the ground truth tokens, and minimize the similarity between the replaced tokens. On the multi-modal encoder, a replace language modeling (RLM) task predicts which token is the replaced one, and a refined image-text matching (RITM) task enforces the image-caption similarity to be higher than the image-rewritten text. 

We demonstrate the effectiveness of the proposed fine-grained VLP framework on various downstream vision-language tasks. The RVLM outperforms weakly-supervised models significantly and is able to achieve compatible or better performance than the fine-grained supervised models.

Our main contributions can be summarized in threefold:
\begin{enumerate}
	\item We present a fine-grained VLP scheme from linguistic perspective, which obtains token-level supervision by the HSR algorithm on image-caption pairs without object annotations.
	\item  A series of refined vision-language modeling tasks, i.e., RITC, RITM, and RLM, are designed to exploit the fine-grained supervision.
	\item Extensive experiments conducted on several downstream vision-language tasks demonstrate the effectiveness of the proposed scheme. As a byproduct, we will release the
	codes, and involved parameters to benefit other researchers.
\end{enumerate}

\begin{figure*}
\centering
	\includegraphics[width=0.85\linewidth]{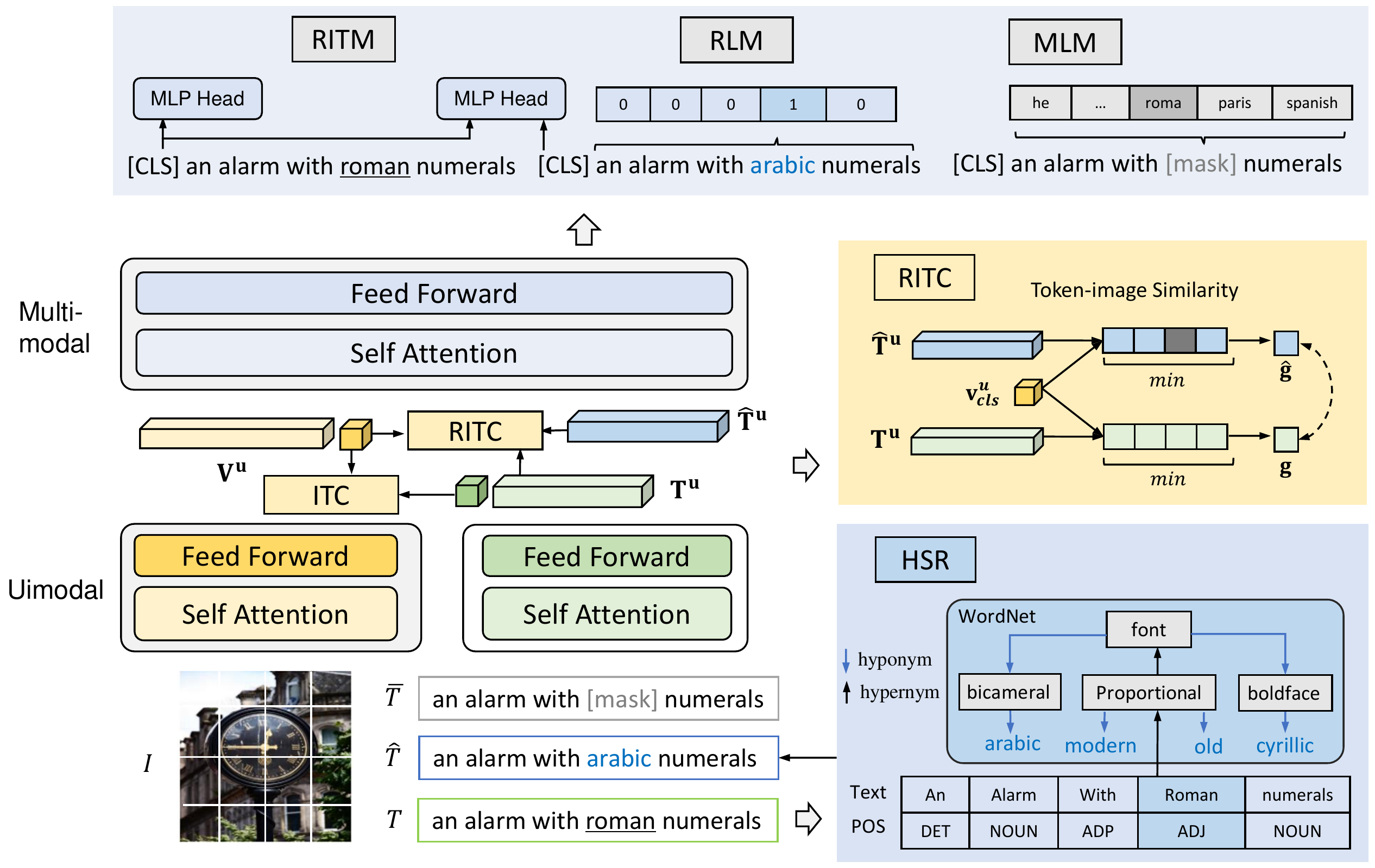}
	\caption{Overall architecture of RVLM. The input caption $T$ is rewritten for negative text $\hat{T}$ by the HSR algorithm. The uni-modal representation is learned through the refined image-text contrastive (RITC) objective. The multi-modal representation is learned through refined image-text matching (RITM) and replace language modeling (RLM) objectives.}
	\label{fig.model}
\end{figure*}

\section{Related Work}

\subsection{Fine-grained VLP}
The object-level annotations used by the early VLP models are from the bottom-up and top-down features~\cite{bottom}, which are produced by a Faster-RCNN~\cite{fasterRCNN:2015} pre-trained on Visual Genome~\cite{VG}. ~\cite{lxmert,vilbert} use two single-modal networks applied for text and images respectively, followed by a cross-modal transformer combining the
two sources. Based on the object annotation, a mask object modeling objective is designed to learn the fine-grained corresponding between the objects and texts. 
The following methods~\cite{visualbert,unicoder,uniter,vlbert,oscar} treat the ROIs as image tokens, and unified the image and text encoder by a BERT~\cite{bert} network. These models improved the pre-training objectives to learn the cross-modal correspondence better and achieve better performance.
VinVL~\cite{vinvl} explores the importance of the object detector and achieves higher performance on downstream vision and language tasks using an improved object detector~\cite{scene}. 
Recently,~\cite{MVPTR} extend the object-tag supervision to multi-grained objects-concepts from scene graph parser~\cite{scene}. Meanwhile, ~\cite{XITM} introduces the object box into the pre-training and proposes a phrase grounding objective. These multi-grained methods achieve significant performance improvement through richer object annotations.
The object detector and scene graph parser are proved to be more compute-expensive~\cite{ALBEF} than ViT~\cite{VIT}. On the other hand, the models rely on expensive bounding box annotations for certain scenes, which inevitably limits VLP's application to more general domains. 
RITM explores fine-grained supervision independent of the object annotations from the language modality. Additionally, RITM is faster than these fine-grained VLP methods because it does not require the object features extraction stage.

\subsection{Weakly-supervised Fine-grained VLP}
The weakly-supervised fine-grained VLP methods do not use object annotations.
~\cite{fashionbert} first explore patch-based image token for VLP in e-commerce retrieval. They split fashion images into patches and use the ResNeXt-101~\cite{resnext} to extract features as the visual input. 
ViLT~\cite{vilt} propose to use the patches tokens as ViT~\cite{VIT}, which is totally convolution-free. Since there are no object annotations, the model is trained with ITM and MLM objectives. The model achieves a higher inference speed than the fine-grained models, but lower performance than the fine-grained VLP methods while using a similar corpus. 
ALBEF~\cite{ALBEF} make up the performance gap by an align before fuse framework. The framework optimizes an ITC objective on the uni-modal encoder output and uses a momentum distillation network to guide the training process.
Based on the framework, ~\cite{li2022blip,coca} propose to use auto-regressive language generation as the language modeling task, and use the captioner to boost the training data scale. These methods achieve excellent performance with the augmented data.  However, it has been reported~\cite{MVPTR} that these methods still fall short of the fine-grained approach when using the same order of magnitude training corpus.
These methods simply use the same language modeling objectives as the uni-modal pre-training BERT~\cite{bert}, and does not fully explore the language modeling tasks for vision and language.
RITM proposes a refined language modeling scheme for VLP, which provides fine-grained supervision from language modality.

\section{Methodology}
The proposed RVLM learns fine-grained cross-modal alignment from image-caption input ${I}$ and ${T}$.
As illustrated in Fig.~\ref{fig.model}, the RVLM consists of three transformer modules: an image encoder and a text encoder for the uni-modal stage, and a cross-modal encoder for the multi-modal stage. 

Formally, we first declare some notations. We use bold capital letters
(e.g., $\mathbf{X}$) to denote the matrices, and bold lowercase letters (e.g., $\mathbf{x}$) for the vectors. The non-bold letters (e.g., $x$) are employed to
represent the scalars and Greek letters (e.g., $\beta$) to denote the parameters.

\subsection{Model Input}
\label{sec.input}
The input image ${I}$ is spitted into 8x8 pixels' patches, and then encoded as sequence of embeddings: $\mathbf{V}=\{\mathbf{v}_{cls}, \mathbf{v}_1, ..., \mathbf{v}_N \}$ following the ViT~\cite{VIT}, where $\mathbf{v}_{cls}$ is the embedding of the [CLS] token. 
The sentence ${T}$ is tokenized as token index sequence $\mathbf{T}=\{\mathbf{w}_{cls}, \mathbf{w}_1, ..., \mathbf{w}_L \}$ where $L$ is the length of the WordPiece~\cite{wordpiece} tokenizer output.  

For the model pre-training, we rewrite the input text ${T}$ as ${\hat{T}}$ through homonyms sentence rewriting for the fine-grained supervision. A masked text $\bar{T}$ is prepared for the MLM.

\subsubsection{Homonyms Sentence Rewriting}
The homonyms sentence rewriting (HSR) aims to provide a fine-grained negative sentence to the image. Based on the principle that token is the language's basic semantic unit, we can change one token of the sentence. The HSR supervision does not require object annotations.

In order to get a hard negative sentence, the semantic of the substitute should be related but different to the ground truth word. Therefore, we propose to choose the homonyms word of the selected token from the WordNet~\cite{wordnet}. Since the captions are descriptive, we only select the concrete noun/adj/verb/quantifier words as the rewriting candidates. 

As shown in the HSR module from the Fig.~\ref{fig.model}, given a sentence $T$, we firstly extract the part-of-speech $S$ of each word, then randomly select one substitute $T_i$ from the noun/adj/verbs/quantifier where $i$ is the word index. Then we search for the hypernyms from WordNet and go through their hyponyms to get the homonyms. If the word does hot have homonyms, we further search for the second-order hypernyms and their second-order hyponyms. The rewritten text $\hat{T}$ is then tokenized to $\mathbf{\hat{T}}$ used for the uni-modal and multi-modal learning.

\subsection{Refined Uni-modal Learning}
\label{sec.uni-modal}
The uni-modal stage takes the $\mathbf{V}$, $\mathbf{T}$ and $\mathbf{\hat{T}}$ as input to learning their fine-grained uni-modal representations.
We use a 6-layer transformer~\cite{transformer} for both the text encoder and a 12-layer visual transformer ViT-B/16~\cite{VIT} as the image encoder. 
The inputs are encoded into the uni-modal representations $\mathbf{V}^u=\{\mathbf{v}^u_{cls}, \mathbf{v}^u_1, ..., \mathbf{v}^u_N \}$, $\mathbf{T}^u=\{\mathbf{w}^u_{cls}, \mathbf{w}^u_1, ..., \mathbf{w}^u_L \}$, and $\hat{\mathbf{T}}^u=\{\mathbf{\hat{w}}^u_{cls}, \mathbf{\hat{w}}^u_1, ..., \mathbf{\hat{w}}^u_L \}$. Afterward, the representation is optimized through ITC and RITC.

\subsubsection{Image-Text Contrastive Learning}
ITC mainly considers coarse-grained alignment between the image and sentence. 
We use the infoNCE~\cite{infonce} loss for the L2-normalized [CLS] representations $h_v(\mathbf{v}^u_{cls})$ and $h_w(\mathbf{w}^u_{cls})$, where $h_v$ and $h_w$ are linear projection layers for the text and image representation. Similar to ~\cite{ALBEF}, a momentum network is used to assist the training. Denote the momentum's representations as $h'_v(\mathbf{v'}_{cls}^u)$ and $h'_w(\mathbf{w'}_{cls}^u)$, and the similarity between text and image as $s(T,I')=h_v(\mathbf{w}^u_{cls})^\intercal h_v(\mathbf{v'}^u_{cls})$,  the text to image contrastive loss can be formulated as Eq.~\ref{equ:contrastive_t}:
\begin{equation}
    \label{equ:contrastive_t}
    \mathcal{L}_{c}^t= 
\end{equation}
where $\tau$ a learnable temperature parameter, $N$ is the batch size and $M$ is the momentum queue length. Symmetrically, we have Eq.~\ref{equ:contrastive_v}.
\begin{equation}
    \label{equ:contrastive_v}
    \mathcal{L}_{c}^t= 
\end{equation}

The ITC loss is calculated using the cross entropy loss on the scores as:
\begin{equation}
    \label{equ.itc}
    \mathcal{L}_{itc} = \mathcal{L}_{c}^t + \mathcal{L}_{c}^v
\end{equation}

\subsubsection{Refined Image-Text Contrastive}
The RITC loss is designed for the fine-grained similarity between the image and tokens. We compute the similarities between image and the tokens $\mathbf{g} =\{h_v(\mathbf{v}_{cls}^u)^\intercal h_w(\mathbf{w}^u_i)\}_{i=1}^L$ and the rewritten text $\hat{\mathbf{g}} =\{{h_v(\mathbf{v}_{cls}^u)}^\intercal h_w(\mathbf{\hat{w}}^u_i)\}_{i=1}^L$. The linear projection layers share the same weights as the ITC.
Afterward, we use the lower bound of $\mathbf{g}$ and $\hat{\mathbf{g}}$ to measure the text-image similarity, so that the fine-grained alignment could be optimized using the margin loss as Eq.~\ref{equ.ritc}:
\begin{equation}
    \label{equ.ritc}
    \mathcal{L}_{ritc} = \max(0, (\beta+\min(\mathbf{g})-\min(\hat{\mathbf{g}})))
\end{equation}
where $\beta$ is the margin hyper parameter. The RITC can be regraded as weakly supervised learning for the alignment between the image and tokens.


\subsection{Refined Multi-modal Learning}
\label{sec.multimodal}
In the multi-modal stage, we enforce cross-modality semantics
learning for coarse-grained alignment with RITM loss, and fine-grained alignment with MLM and RLM loss. 
The multi-modal encoder is a 6-layer cross-transformer that fuses the uni-modal representations into the language modality. It produces the multi-modal representation $\mathbf{T}^m$, $\bar{\mathbf{T}}^m$ and $\hat{\mathbf{T}}^m$ for the caption, masked and rewritten texts.

\subsubsection{Refined Image-Text Matching}
The RITM loss learns a global representation of the text-image pair to predict if the text and image match. We employ two kinds of negative pairs for the matching. The first is the in-batch hard negative samples, which are selected according to the similarity scores from the uni-modal stage. The second is the image and rewritten text.
The loss function is a two-fold binary classification loss on the [CLS] representation, which is calculated as Eq.~\ref{equ.ritm}.
\begin{equation}
    \label{equ.ritm}
    \begin{aligned}
    \mathcal{L}_{ritm} = \mathbb{E}_{(I,T,\hat{T}) \sim D}& H(\mathbf{y}^{itm},\mathbf{p}^{ritm}(\mathbf{T}^m_{cls})) \\
    &+H(\mathbf{\hat{y}}^{itm},\mathbf{\hat{p}}^{ritm}(\mathbf{\hat{T}}^m_{cls}))
    \end{aligned}
\end{equation}
where $\mathbf{y}^{itm}$ and $\mathbf{\hat{y}}^{itm}$ are 2-dimensional one-hot vectors representation the alignment label, $H$ denotes the cross entropy function, $\mathbf{p}^{ritm}$ and $\mathbf{\hat{p}}^{ritm}$ are one-layer linear projections with softmax that project the [CLS] representations into binary probability.

\subsubsection{Mask Language Modeling}
The MLM with visual clues is traditional and important for VLP. It utilizes multi-modal context to predict the masked words.
In addition to the traditional setting that randomly masks out $15\%$ of the input tokens, we also mask the replaced token indicated by the HSR. Afterward, $80\%$ masked tokens are replaced by [MASK] and $10\%$ are replaced by random tokens. 
The MLM term is formulate as Eq.~\ref{equ.mlm}
\begin{equation}
    \label{equ.mlm}
    \mathcal{L}_{mlm} = \mathbb{E}_{(I,\bar{T}) \sim D}H(\mathbf{y}^{msk},\mathbf{p}^{msk}(\bar{\mathbf{T}}^m))
\end{equation}
where $\mathbf{y}^{msk}$ is a one-hot vocabulary distribution where the ground-truth token has a probability of 1, $\mathbf{p}^{mlm}$ is the model's prediction head network.

MLM objective learns fine-grained correspondence between the multi-modal context and the token through classification. There are two disadvantages: First, there may be many proper tokens to fill the mask, while the objective ignores the other correction tokens. Second, it does not directly learn the alignment between the multi-modal context and the token.

\begin{table}[] 
\centering
\setlength{\tabcolsep}{3pt}
\begin{tabular}{c|cccccc|c}
    \hline
    & COCO & VG   & CC   & SBU  & OI  &Flickr30k & Toal\\ \hline
     image & 113k & 100k & 2.8M & 860k & 504k & 30k & 4.4M\\
     text  & 567k & 769k & 2.8M & 860k & 507k & 150k & 5.7M\\ \hline
\end{tabular}
\caption{Statistics of the pre-training datasets.}
\label{tab.dataset}
\end{table}

\begin{table*}
\centering
\small
\setlength{\tabcolsep}{3pt}
\begin{tabular}{lccclcccllccclccclc} 
	\toprule
	\multirow{3}{*}{Methods} & \multicolumn{8}{c}{Flickr30k Test (1k images)} &  & \multicolumn{7}{c}{COCO Test (5k images)} & {} & \multirow{2}{*}{$\#$ Pre-train}\\ 
	\cmidrule{2-8} \cmidrule{11-17}
	& \multicolumn{3}{c}{Text Retrieval} & {} & \multicolumn{3}{c}{Image Retrieval} &  &  & \multicolumn{3}{c}{Text Retrieval} & {} & \multicolumn{3}{c}{Image Retrieval} &  & \multirow{1}{*}{Images} \\ 
	\cmidrule{2-4}\cmidrule{6-8}\cmidrule{11-13}\cmidrule{15-17}
	& R@1 & R@5 & R@10 &  & R@1 & R@5 & R@10 & AR &  & R@1 & R@5 & R@10 &  & R@1 & R@5 & R@10 & AR \\ 
	\midrule
	\multicolumn{17}{c}{Weakly-supervised Fine-grained VLP} \\
	ALIGN~\shortcite{ALIGN} & 95.3 & 99.8 & 100 &  & 84.9 & 97.4 & 98.6 & 96.0 &  & 77.0 & 93.5 & 96.9 & & 59.9 & 83.3 & 89.8 & 83.4 & 1.8B \\ 
	FILIP~\shortcite{filip} & \textbf{96.6} & \textbf{100} & \textbf{100} &  & \textbf{87.1} & \textbf{97.7} & \textbf{99.1} & \textbf{96.8} &  & 78.9 & 94.4 & 97.4 &  &  61.2 & 84.3 & 90.5 & 84.5 & 340M\\
	\cmidrule{2-19}
	ViLT~\shortcite{vilt} & 83.5 & 96.7 & 98.6 &  & 64.4 & 88.7 & 93.8 & 87.6 &  &  61.5 & 86.3 & 92.7 & & 42.7 & 72.9 & 83.1 & 73.2 & 14M \\ 
	ALBEF~\shortcite{ALBEF} & 94.3 & 99.4 & 99.8 &  & 82.8 & 96.7 & 98.4 & 95.2 &  & 73.1 & 91.4 & 96.0 & & 56.8 & 81.5 & 89.2 & 81.3 & 4M \\ 
	ALBEF-14M~\shortcite{ALBEF} & 95.9 & 99.8 & 100 &  & 85.6 & 97.5 & 98.9 & 96.2 &  & 77.6 & 94.3 & 97.2 & & 60.7 & 84.3 & 90.5 & 84.1 & 14M \\ 
	\midrule
	\multicolumn{17}{c}{ Fine-grained Supervised VLP} \\
	Uniter~\shortcite{uniter} & 85.9 & 97.1 & 98.8 &  & 72.5 & 92.4 & 96.1 & 90.5 &  & 65.7 & 88.6 & 93.8 &  & 52.9 & 79.9 & 88.0 & 78.1 & 4M\\
	Oscar~\shortcite{oscar} & - & - & - &  & - & - & - & - &  & 70.0 & 91.1 & 95.5 &  &  54.0 & 80.8 & 88.5 & 80.0 & 4M\\
	UNIMO~\shortcite{unimo} & 89.7 & 98.4 & 99.1 & & 74.7 & 93.4 & 99.1 & 92.4 &  & - & - & - &  & - & - & - & - & 4M \\ 
	ROSITA~\shortcite{rosita} & 88.9 & 98.1 & 99.3 & & 74.1 & 92.4 & 96.1 & 91.4 &  & 71.3 & 91.6 & 95.6 &  & 54.4 & 80.9 & 88.6 & 80.4 & 4M \\ 
	VinVL~\shortcite{vinvl} & 93.6 & 99.1 & 99.9 &  & 82.0 & 95.7 & 97.7 & 94.6 &  & 74.6 & 92.6 & 96.3 &  & 58.1 & 83.2 & 90.1 & 82.5 & 5.7M\\ 
	MVPTR~\shortcite{MVPTR} & 95.2 & 99.7 & 100 &  & 84.0 & 96.8 & 98.5 & 95.7 &  & 77.3 & 93.6 & 96.9 & & 60.1 & 84.0 & 90.7& 83.8 & 4.7M\\ \midrule
	RVLM & 94.7 & 99.8 & 99.8 &  & 82.4 & 96.1 & 98.0 & 95.1 &  & 77.3 & 94.2 & 97.2 &  & 60.9 & 84.6 & 91.2 & 84.2 & 4.4M\\ 
	RVLM-Large & 95.6 & {99.8} & \textbf{100} &  & {85.7} & {97.6} & 98.8 & {96.2} &  & \textbf{79.5} & \textbf{95.1} & \textbf{97.9} &  & \textbf{63.1} & \textbf{85.6} & \textbf{91.9} & \textbf{85.5} & 14M\\ 
	\bottomrule
\end{tabular}
\caption{Image-text retrieval results on Flickr30K and COCO datasets in R@$k$ metrics.}
\label{tab.retrieval}
\end{table*}

\subsubsection{Replace Language Modeling}
The replace language modeling task predicts which token is replaced according to the multi-modal content. It makes up the two above-mentioned limitations of MLM. Firstly, the binary token alignment is certain. Secondly, the objective directly learns the fine-grained alignment between the image and tokens. The fine-grained alignment learning is weakly-supervised. The calculation of the loss is formulated as Eq.~\ref{equ.rlm}
\begin{equation}
    \label{equ.rlm}
    \mathcal{L}_{rlm} = \mathbb{E}_{(I,\hat{T}) \sim D}H(\mathbf{y}^{rlm},\mathbf{p}^{rlm}(\hat{\mathbf{T}}^m))
\end{equation}
where $\mathbf{y}^{rlm}$ is a one-hot distribution of length $L$, where the replaced token has a probability of 1, $\mathbf{p}^{rlm}$ is the model's prediction head network. 

The full pre-training objective of RVLM is the combination of the above motioned terms:
\begin{equation}
    \label{equ.RVLM}
    \mathcal{L} = \mathcal{L}_{itc} + \mathcal{L}_{ritc} + \mathcal{L}_{ritm} + \mathcal{L}_{mlm} + \mathcal{L}_{rmlm}
\end{equation}
The model is optimized end-to-end on the pre-training datasets to minimize $\mathcal{L}$.

\subsection{Pre-training Datasets}

We pre-train our model on a large-scale vision-language corpus, including
MSCOCO~\cite{coco}, Visual Genome~\cite{VG}, Flickr30k~\cite{flickr30k}, Conceptual Captions~\cite{cc}\footnote{$\sim$0.5M images are excluded for bad URLs.}, SBU~\cite{sbu} and OpenImages~\cite{openimages}. Table.~\ref{tab.dataset} shows the statistics of the image and text of the pre-training datasets. 

We exclude the val/test splits of Flickr30K as ~\cite{MVPTR}. Additionally, we scale up RVLM as ~\cite{ALBEF} using a larger-scale web data Conceptual Concept 12M~\cite{cc12M}, which totally uses 14M images.
The large SBU and Conceptual Captions datasets are reported~\cite{hendricks} to be noisy, where the captions may not correspond exactly to the image. Therefore, the token-level supervision of HSR may be blurred by the noisy words. To solve the problem, we divide the training process into two stages: the first 30 epochs are trained on all pre-training corpus, and then 10 epochs on the two human-annotated MSCOCO and Visual Genome datasets.  

\subsection{Implementation Details}
\label{sec.implement}
The image encoder is initialized with the first 12 layers of MAE-base~\cite{mae} weight, which is pre-trained on the ImageNet-1k without any labels. The text encoder and cross-modal encoder are initialized with the BERT-base~\cite{bert} weights.
The AdamW~\cite{adamw} optimizer is adopted with a learning rate of $1e^{-4}$ with $0.02$ weight decay, and warm-up from the first 20 epochs, then linearly decay 10 epochs to 0. The batch size is set to $32$ with momentum queue size $65536$. 
The max sequence length of text tokens is set to $30$. The model is trained with half-precision on NVIDIA DGX with Ubuntu system and 8 V100 GPU. 

\section{Experiments}
To evaluate the proposed method, we conducted extensive experiments on three kinds of downstream V+L tasks.  We introduce each task and the experimental results below.

\subsection{Image-text Retrieval}
We evaluate RITM for both image-to-text retrieval and text-to-image retrieval. The benchmarks are Flickr30K and COCO. The retrieval performance is measured by the recall at top-k samples (R@k). Three $k$ values, R@$1$, R@$5$, and R@$10$, are reported for text-to-image retrieval and vice versa. The fine-tuning is optimized with $\mathcal{L}_u+\mathcal{L}_{ritm}$ using the same set of pre-training. 
During inference, we employ the recall and re-ranking strategy as ~\cite{ALBEF} to improve retrieval efficiency. We first recall top-$k$ candidates using the uni-modal similarity score as Eq.~\ref{equ.itc} for all image-text pairs, then calculate their ITM score for ranking. In addition, we also include the CXC~\cite{CXC} metrics, which are reported to provide a more precise ranking for the COCO images.

Table~\ref{tab.retrieval} shows the image-text retrieval comparison between the RVLM  and state-of-the-art VLP methods on COCO 5k and Flickr30k 1k test sets. Overall, the object annotation supervised methods perform better than the weakly-supervised methods. However, RVLM achieves comparable performance to object annotation supervised methods like MVPTR. Compared with the weakly-supervised methods, RVLM achieves higher performance, especially on the COCO dataset. On the 4M pre-training setting, RVLM outperforms ALBEF by $4.2\%$ and $4.1\%$ in terms of the R@$1$ score, indicating that RVLM learns better fine-grained alignments to distinguish similar images. Compared with the object annotation supervised MVPTR model, RVLM has a small $0.6\%$ gap in average to MVPTR on Flickr30k but outperforms it by $0.5\%$ on the COCO 5k test set.

Since some queries have multiple semantically aligned images or text~\cite{CXC}, we further conduct an experiment on the CXC metrics to verify our improvement on the COCO dataset. As the results in Table~\ref{tab:cxc}, RVLM still constantly outperforms the compared methods and achieves the highest improvement on the R@$1$ score, which qualifies that RVLM is good at learning fine-grained alignments. 
\begin{table}
\setlength{\tabcolsep}{3pt}
\centering  
\begin{tabular}{lllllllll} 
\toprule
\multirow{3}{*}{Methods} & \multicolumn{8}{c}{COCO Test (CXC)} \\ 
\cmidrule{2-9}
 & \multicolumn{3}{c}{Text Retrieval} & {} & \multicolumn{3}{c}{Image Retrieval} &  \\ 
\cmidrule{2-4}\cmidrule{6-8}
 & {R@1} & {R@5} & {R@10} &  & {R@1} & {R@5} & {R@10}  \\ 
\midrule
ALBEF     & 74.48 & 92.34 & 96.20 &  & 59.43 & 83.67 & 89.87\\
ALBEF-14M & 78.48 & 94.66 & 97.64 &  & 63.63 & 85.80 & 91.45\\
MVPTR     & 78.16 & 94.98 & 97.39 &  & 63.52 & 85.03 & 90.97\\ 
\midrule
RVLM     & 78.22 & 95.10 & 97.70 &  & 63.74 & 86.46 & 92.22  \\
RVLM-14M & \textbf{80.76} & \textbf{95.32} & \textbf{98.16} &  & \textbf{65.64 }& \textbf{87.18} & \textbf{92.62} \\
\bottomrule
\end{tabular}
\caption{Image-text retrieval results on COCO dataset with CXC~\cite{CXC} metrics.}
\label{tab:cxc}
\vspace{-5pt}
\end{table}


\subsection{Multi-Modal Classification}
We evaluate RVLM on two widely-used multi-modal classification tasks, i.e., visual question answering (VQA) and visual entailment (VE).

VQA requires the model to select an answer given an image and a question as context. We use the commonly used VQA v2~\cite{vqav2} benchmark as the evaluation dataset. The evaluation metric is the accuracy on the answer selection. Following~\cite{ALBEF,cho2021unifying}, we consider VQA as an answer generation problem. The RVLM takes the image and question as input and generates $3,192$ candidate answers using a 6-layer transformer decoder.

VE is a fine-grained visual reasoning task to predict the relationship between an image and a text. The relationship categories are entailment, neutral, or contradictory. The evaluation metric is the accuracy of the relation classification. We follow the common practice of VLP~\cite{ALBEF,uniter} to formulate VE as a three-way classification problem, and predict the class probabilities using a multi-layer perceptron (MLP) layer on the [CLS] token representation $\mathbf{T}^m_{cls}$ of the multimodal encoder.

\begin{table}
\centering  
\begin{tabular}{llllllll} 
\toprule
\multirow{2}{*}{Methods} & \multicolumn{2}{c}{VQA} &  & \multicolumn{2}{c}{SNLI-VE} \\ 
\cmidrule{2-3}\cmidrule{5-6}
 & test-dev & test-std &  & val & test \\ 
\midrule
ViLT$^\ddagger$   & 70.94 & -     &  & -     & - \\
ALBEF$^\ddagger$  & 74.54 & 74.70 &  & 80.14 & 80.30 \\ \midrule
Uniter  & 72.70 & 72.91 &  & 78.59 & 78.28 \\
OSCAR  & 73.16 & 73.44 &  & -     & - \\
VILLA  & 73.59 & 73.67 &  & 79.47 & 79.03 \\
UNIMO  & 73.79 & 74.02 &  & 80.00 & 79.10 \\
VinVL  & 75.95 & 76.12 &  & -     & - \\
MVPTR  & \textbf{76.16} & \textbf{76.36} &  & 80.30 & 80.17 \\
\midrule
RVLM$^\ddagger$   & 75.85 & 75.93 &  & \textbf{80.52} & \textbf{80.41} \\ 
\bottomrule
\end{tabular}
\caption{Multi-modal classification results on VQA v2 and SNLI-VE datasets (Models with $^\ddagger$ are weakly-supervised methods). }
\label{tab:classification}
\end{table}
Table~\ref{tab:classification} shows the VQA and VE comparison between the RVLM and state-of-the-art VLP models. On both of the two tasks, object annotation based models are sightly better than existing weakly-supervised models. On the VQA v2 benchmark, RVLM outperforms other weakly-supervised methods and achieves close accuracy than the state-of-the-art object annotation supervised models, proving RVLM's effectiveness in learning fine-grained semantics.  On the VE task, RVLM outperforms both of the two kinds of methods, proving the model's superior ability to learn cross-modal relationships.

\subsection{Weakly-supervised Visual Grounding}
\begin{table}[]
    \centering
    \begin{tabular}{lccc}
        \toprule
         \multirow{2}{*}{Method} & \multicolumn{3}{c}{RefCOCO+}\\ \cmidrule{2-4}
         & dev & testA & testB \\ \midrule
         ARN~\cite{arn} & 32.78 & 34.35 & 32.13 \\
         CCL~\cite{ccl} & 34.29 & 36.91 & 33.56 \\
         ALBEF-14M  & 58.46 & 65.89 & 46.25\\ \midrule
         RVLM-14M  & \textbf{58.75} & \textbf{67.32} & \textbf{46.74} \\ \bottomrule
    \end{tabular}
    \caption{Weakly-supervised visual grounding results on RefCOCO+ dataset.}
    \label{tab:grounding}
    \vspace{-5pt}
\end{table}
Visual grounding requires models to localize the region of the specific textual description from the input image. Our RVLM is fine-tuned on the RefCOCO+~\cite{refcoco+} dataset.
Since RVLM does not rely on any object box inputs, it can not use the bounding box annotations of the dataset as supervision.
Therefore, we use the weakly-supervised setting following~\cite{ALBEF} and fine-tune the 14M pre-trained model on the image-text matching score $\mathcal{L}_{itc}$. During inference, we firstly obtain heatmaps on each pixel using Grad-CAM~\cite{gradcam}, and use them to rank the detected proposals provided by~\cite{yu2018mattnet}. 
Table~\ref{tab:grounding} shows the weakly-supervised visual grounding comparison results on RefCOCO+. RVLM outperforms existing weakly-supervised methods on both the two test sets, proving its ability in learning fine-grained alignments. 

\begin{figure*}
    \centering
\includegraphics[width=\linewidth]{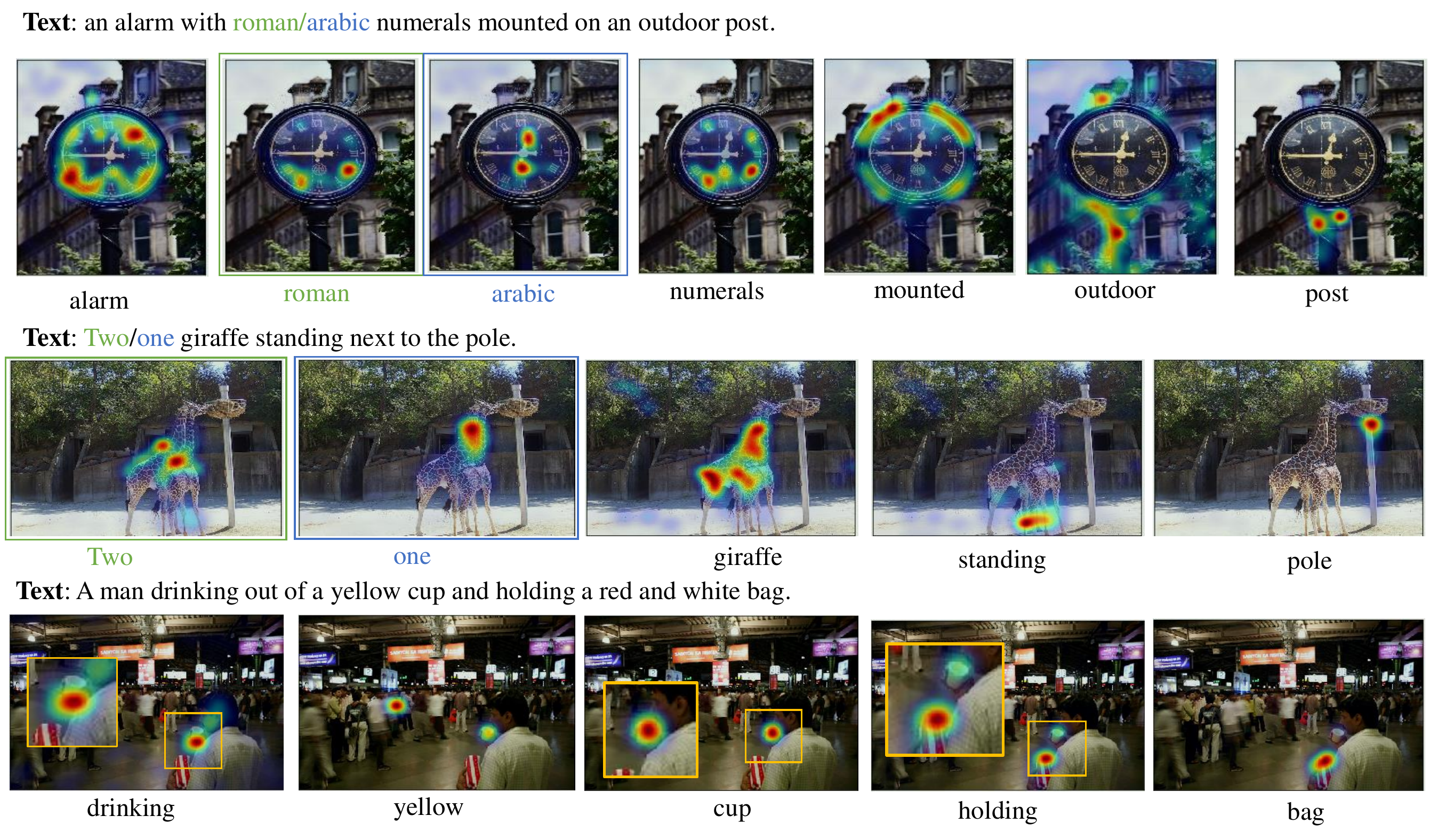} 
    \caption{Visualization of the Grad-CAM on cross-attention maps between region and words (Better viewed with zoom-in). The images surrounded by green or blue boxes are from the sentence with corresponding words.}
    \label{fig:visualization}
    \vspace{-5pt}
\end{figure*}

\subsection{Further Analysis}
\subsubsection{Fine-grained Alignment Visualization by Grad-CAM}
To inspect the token-level alignment learned by RVLM, we visualize the cross-attention map on the image for each word using Grad-CAM. A few visualizations are provided in Fig.~\ref{fig:visualization}. In the first row, both the nouns (e.g., alarm and post) and verbs (mount) are aligned to their corresponding regions. Specially, we show the attention map of the substitute word ``roman'' and ``arabic'' for the same content. RVLM aligns ``roman'' correctly, and the  ''arabic'' was not aligned to the roman numbers. The second row includes the quantifier words ``one'' and ``two'', which are aligned to correct number of regions on the giraffes. The third row shows the fine-grained alignment between the image and multiple objects. It can be observed that the tokens are precisely aligned to the objects. Specifically, the two verbs, ``drink'' is correctly connected to the cup and mouth, and ``hold'' to both hands, proving that RVLM is able to distinguish the verbs. The two nuns, ``cup'' and ``bag'' are also correctly located, verifying RVLM's fine-grained alignment capability.

\begin{table}
\centering \small
\begin{tabular}{lllllllll} 
\toprule
\multirow{2}{*}{Methods} & \multicolumn{2}{c}{SNLI-VE} &  & \multicolumn{2}{c}{COCO (5k)} &  \\ 
\cmidrule{2-3}\cmidrule{5-7}
 & dev & test &  & I2T & T2I \\ 
\midrule
RVLM      & 80.52 & 80.41 &    & 77.3 & 60.9 & \\ 
\midrule
w/o RITM  & 80.32 & 80.23 &  & 77.0 & 59.7 &  \\
w/o RITC   & 80.14 & 80.11 &  & 75.9 & 59.3 &  \\
w/o RLM  & 79.09 & 79.03 &  & 76.7 & 58.9 &  \\
w/o HSR   & 78.89 & 78.24 & & 75.1 & 58.8 &  \\
\bottomrule
\end{tabular}
\caption{Ablation study on the VE and image-text retrieval task.  R@$1$ score is used on COCO 5k test set.}
\label{tab:ablation}
\vspace{-5pt}
\end{table}
\subsubsection{Ablation study}
We conduct detailed ablation studies on SNLI-VE and COCO 5k datasets to analyze the effectiveness of each proposed component. Specifically, we remove the RITM, RLM, and RITC loss in the second stage of pre-training. We also verify the HSR strategy by using randomly selected words for rewriting. The variants are trained using the same hyper-parameters and epoch.
Table~\ref{tab:ablation} shows the results of the two benchmarks. It could be observed that the removal of each component results in a performance decrease for both tasks. Using randomly selected words for rewriting results in a significant performance decrease, proving the effectiveness of homonyms words.

\section{Conclusion}
In this paper, we proposed a novel refined vision-language modeling (RVLM) for weakly-supervised fine-grained vision-language pre-training. Instead of using object annotations, we proposed a homonyms sentence rewriting (HSR) algorithm to produce a token-level negative sentence. To capture the fine-grained supervision from the rewritten sentence, a refined image-text contrastive loss (RITC) is proposed to learn the fine-grained alignment at the uni-modal stage, and refined image-text matching (RITM) and replace language modeling (RLM) task at the multi-modal stage. Extensive experiments and analysis showed that RVLM outperformed state-of-the-art models on several downstream benchmarks.

Several future directions and improvements could be considered.
Our main objective was to show the potential of the weakly-supervised refined vision-language modeling framework. 
The HSR only provides token-level supervision, a phrase-level rewriting strategy could be considered to develop phrase-level supervision.
Besides, developing the rewriting technique for image modality is also promising for fine-grained supervision.

\bibliography{aaai23}

\begin{thebibliography}{48}
\providecommand{\natexlab}[1]{#1}

\bibitem[{Anderson et~al.(2016)Anderson, Fernando, Johnson, and Gould}]{scene}
Anderson, P.; Fernando, B.; Johnson, M.; and Gould, S. 2016.
\newblock Spice: Semantic propositional image caption evaluation.
\newblock In \emph{European conference on computer vision}, 382--398. Springer.

\bibitem[{Anderson et~al.(2018)Anderson, He, Buehler, Teney, Johnson, Gould,
  and Zhang}]{bottom}
Anderson, P.; He, X.; Buehler, C.; Teney, D.; Johnson, M.; Gould, S.; and
  Zhang, L. 2018.
\newblock Bottom-up and top-down attention for image captioning and visual
  question answering.
\newblock In \emph{Proceedings of the IEEE conference on computer vision and
  pattern recognition}, 6077--6086.

\bibitem[{Changpinyo et~al.(2021)Changpinyo, Sharma, Ding, and Soricut}]{cc12M}
Changpinyo, S.; Sharma, P.; Ding, N.; and Soricut, R. 2021.
\newblock Conceptual 12m: Pushing web-scale image-text pre-training to
  recognize long-tail visual concepts.
\newblock In \emph{Proceedings of the IEEE/CVF Conference on Computer Vision
  and Pattern Recognition}, 3558--3568.

\bibitem[{Chen et~al.(2020)Chen, Li, Yu, El~Kholy, Ahmed, Gan, Cheng, and
  Liu}]{uniter}
Chen, Y.-C.; Li, L.; Yu, L.; El~Kholy, A.; Ahmed, F.; Gan, Z.; Cheng, Y.; and
  Liu, J. 2020.
\newblock Uniter: Universal image-text representation learning.
\newblock In \emph{European conference on computer vision}, 104--120.

\bibitem[{Cho et~al.(2021)Cho, Lei, Tan, and Bansal}]{cho2021unifying}
Cho, J.; Lei, J.; Tan, H.; and Bansal, M. 2021.
\newblock Unifying vision-and-language tasks via text generation.
\newblock In \emph{International Conference on Machine Learning}, 1931--1942.
  PMLR.

\bibitem[{Cui et~al.(2021)Cui, Yu, Wang, Zhao, Zhang, Wang, and Yu}]{rosita}
Cui, Y.; Yu, Z.; Wang, C.; Zhao, Z.; Zhang, J.; Wang, M.; and Yu, J. 2021.
\newblock ROSITA: Enhancing Vision-and-Language Semantic Alignments via
  Cross-and Intra-modal Knowledge Integration.
\newblock In \emph{Proceedings of the 29th ACM International Conference on
  Multimedia}, 797--806.

\bibitem[{Devlin et~al.(2019)Devlin, Chang, Lee, and Toutanova}]{bert}
Devlin, J.; Chang, M.-W.; Lee, K.; and Toutanova, K. 2019.
\newblock BERT: Pre-training of Deep Bidirectional Transformers for Language
  Understanding.
\newblock In \emph{Proceedings of the 2019 Conference of the North American
  Chapter of the Association for Computational Linguistics: Human Language
  Technologies, Volume 1 (Long and Short Papers)}, 4171--4186.

\bibitem[{Dosovitskiy et~al.(2020)Dosovitskiy, Beyer, Kolesnikov, Weissenborn,
  Zhai, Unterthiner, Dehghani, Minderer, Heigold, Gelly et~al.}]{VIT}
Dosovitskiy, A.; Beyer, L.; Kolesnikov, A.; Weissenborn, D.; Zhai, X.;
  Unterthiner, T.; Dehghani, M.; Minderer, M.; Heigold, G.; Gelly, S.; et~al.
  2020.
\newblock An Image is Worth 16x16 Words: Transformers for Image Recognition at
  Scale.
\newblock In \emph{International Conference on Learning Representations}.

\bibitem[{Gao et~al.(2020)Gao, Jin, Chen, Qiu, Li, Wei, Hu, and
  Wang}]{fashionbert}
Gao, D.; Jin, L.; Chen, B.; Qiu, M.; Li, P.; Wei, Y.; Hu, Y.; and Wang, H.
  2020.
\newblock Fashionbert: Text and image matching with adaptive loss for
  cross-modal retrieval.
\newblock In \emph{Proceedings of the 43rd International ACM SIGIR Conference
  on Research and Development in Information Retrieval}, 2251--2260.

\bibitem[{Goyal et~al.(2017)Goyal, Khot, Summers-Stay, Batra, and
  Parikh}]{vqav2}
Goyal, Y.; Khot, T.; Summers-Stay, D.; Batra, D.; and Parikh, D. 2017.
\newblock Making the v in vqa matter: Elevating the role of image understanding
  in visual question answering.
\newblock In \emph{Proceedings of the IEEE conference on computer vision and
  pattern recognition}, 6904--6913.

\bibitem[{He et~al.(2022)He, Chen, Xie, Li, Doll{\'a}r, and Girshick}]{mae}
He, K.; Chen, X.; Xie, S.; Li, Y.; Doll{\'a}r, P.; and Girshick, R. 2022.
\newblock Masked autoencoders are scalable vision learners.
\newblock In \emph{Proceedings of the IEEE/CVF Conference on Computer Vision
  and Pattern Recognition}, 16000--16009.

\bibitem[{Hendricks et~al.(2021)Hendricks, Mellor, Schneider, Alayrac, and
  Nematzadeh}]{hendricks}
Hendricks, L.~A.; Mellor, J.; Schneider, R.; Alayrac, J.-B.; and Nematzadeh, A.
  2021.
\newblock Decoupling the role of data, attention, and losses in multimodal
  transformers.
\newblock \emph{Transactions of the Association for Computational Linguistics},
  9: 570--585.

\bibitem[{Jia et~al.(2021)Jia, Yang, Xia, Chen, Parekh, Pham, Le, Sung, Li, and
  Duerig}]{ALIGN}
Jia, C.; Yang, Y.; Xia, Y.; Chen, Y.-T.; Parekh, Z.; Pham, H.; Le, Q.~V.; Sung,
  Y.; Li, Z.; and Duerig, T. 2021.
\newblock Scaling up visual and vision-language representation learning with
  noisy text supervision.
\newblock \emph{arXiv preprint arXiv:2102.05918}.

\bibitem[{Johnson et~al.(2017)Johnson, Schuster, Le, Krikun, Wu, Chen, Thorat,
  Vi{\'e}gas, Wattenberg, Corrado et~al.}]{wordpiece}
Johnson, M.; Schuster, M.; Le, Q.~V.; Krikun, M.; Wu, Y.; Chen, Z.; Thorat, N.;
  Vi{\'e}gas, F.; Wattenberg, M.; Corrado, G.; et~al. 2017.
\newblock Google’s multilingual neural machine translation system: Enabling
  zero-shot translation.
\newblock \emph{Transactions of the Association for Computational Linguistics},
  5: 339--351.

\bibitem[{Kim, Son, and Kim(2021)}]{vilt}
Kim, W.; Son, B.; and Kim, I. 2021.
\newblock Vilt: Vision-and-language transformer without convolution or region
  supervision.
\newblock In \emph{International Conference on Machine Learning}, 5583--5594.
  PMLR.

\bibitem[{Krishna et~al.(2017)Krishna, Zhu, Groth, Johnson, Hata, Kravitz,
  Chen, Kalantidis, Li, Shamma et~al.}]{VG}
Krishna, R.; Zhu, Y.; Groth, O.; Johnson, J.; Hata, K.; Kravitz, J.; Chen, S.;
  Kalantidis, Y.; Li, L.-J.; Shamma, D.~A.; et~al. 2017.
\newblock Visual genome: Connecting language and vision using crowdsourced
  dense image annotations.
\newblock \emph{International journal of computer vision}, 123(1): 32--73.

\bibitem[{Kuznetsova et~al.(2020)Kuznetsova, Rom, Alldrin, Uijlings, Krasin,
  Pont-Tuset, Kamali, Popov, Malloci, Kolesnikov et~al.}]{openimages}
Kuznetsova, A.; Rom, H.; Alldrin, N.; Uijlings, J.; Krasin, I.; Pont-Tuset, J.;
  Kamali, S.; Popov, S.; Malloci, M.; Kolesnikov, A.; et~al. 2020.
\newblock The open images dataset v4.
\newblock \emph{International Journal of Computer Vision}, 128(7): 1956--1981.

\bibitem[{Li et~al.(2020{\natexlab{a}})Li, Duan, Fang, Gong, and
  Jiang}]{unicoder}
Li, G.; Duan, N.; Fang, Y.; Gong, M.; and Jiang, D. 2020{\natexlab{a}}.
\newblock Unicoder-vl: A universal encoder for vision and language by
  cross-modal pre-training.
\newblock In \emph{Proceedings of the AAAI Conference on Artificial
  Intelligence}, volume~34, 11336--11344.

\bibitem[{Li et~al.(2022{\natexlab{a}})Li, Li, Xiong, and Hoi}]{li2022blip}
Li, J.; Li, D.; Xiong, C.; and Hoi, S. 2022{\natexlab{a}}.
\newblock Blip: Bootstrapping language-image pre-training for unified
  vision-language understanding and generation.
\newblock \emph{arXiv preprint arXiv:2201.12086}.

\bibitem[{Li et~al.(2021{\natexlab{a}})Li, Selvaraju, Gotmare, Joty, Xiong, and
  Hoi}]{ALBEF}
Li, J.; Selvaraju, R.; Gotmare, A.; Joty, S.; Xiong, C.; and Hoi, S. C.~H.
  2021{\natexlab{a}}.
\newblock Align before fuse: Vision and language representation learning with
  momentum distillation.
\newblock volume~34.

\bibitem[{Li et~al.(2019)Li, Yatskar, Yin, Hsieh, and Chang}]{visualbert}
Li, L.~H.; Yatskar, M.; Yin, D.; Hsieh, C.-J.; and Chang, K.-W. 2019.
\newblock VisualBERT: A Simple And Performant Baseline For Vision And Language.
\newblock \emph{arXiv preprint arXiv:1908.03557}.

\bibitem[{Li et~al.(2021{\natexlab{b}})Li, Gao, Niu, Xiao, Liu, Liu, Wu, and
  Wang}]{unimo}
Li, W.; Gao, C.; Niu, G.; Xiao, X.; Liu, H.; Liu, J.; Wu, H.; and Wang, H.
  2021{\natexlab{b}}.
\newblock UNIMO: Towards Unified-Modal Understanding and Generation via
  Cross-Modal Contrastive Learning.
\newblock In \emph{Proceedings of the 59th Annual Meeting of the Association
  for Computational Linguistics and the 11th International Joint Conference on
  Natural Language Processing (Volume 1: Long Papers)}, 2592--2607.

\bibitem[{Li et~al.(2020{\natexlab{b}})Li, Yin, Li, Zhang, Hu, Zhang, Wang, Hu,
  Dong, Wei et~al.}]{oscar}
Li, X.; Yin, X.; Li, C.; Zhang, P.; Hu, X.; Zhang, L.; Wang, L.; Hu, H.; Dong,
  L.; Wei, F.; et~al. 2020{\natexlab{b}}.
\newblock Oscar: Object-semantics aligned pre-training for vision-language
  tasks.
\newblock In \emph{European Conference on Computer Vision}, 121--137.

\bibitem[{Li et~al.(2022{\natexlab{b}})Li, Fan, Tou, and Wei}]{MVPTR}
Li, Z.; Fan, Z.; Tou, H.; and Wei, Z. 2022{\natexlab{b}}.
\newblock MVP: Multi-Stage Vision-Language Pre-Training via Multi-Level
  Semantic Alignment.
\newblock \emph{arXiv preprint arXiv:2201.12596}.

\bibitem[{Lin et~al.(2014)Lin, Maire, Belongie, Hays, Perona, Ramanan,
  Doll{\'a}r, and Zitnick}]{coco}
Lin, T.-Y.; Maire, M.; Belongie, S.; Hays, J.; Perona, P.; Ramanan, D.;
  Doll{\'a}r, P.; and Zitnick, C.~L. 2014.
\newblock Microsoft coco: Common objects in context.
\newblock In \emph{European conference on computer vision}, 740--755. Springer.

\bibitem[{Liu et~al.(2019)Liu, Li, Wang, Zha, Meng, and Huang}]{arn}
Liu, X.; Li, L.; Wang, S.; Zha, Z.-J.; Meng, D.; and Huang, Q. 2019.
\newblock Adaptive reconstruction network for weakly supervised referring
  expression grounding.
\newblock In \emph{Proceedings of the IEEE/CVF International Conference on
  Computer Vision}, 2611--2620.

\bibitem[{Loshchilov and Hutter(2018)}]{adamw}
Loshchilov, I.; and Hutter, F. 2018.
\newblock Decoupled Weight Decay Regularization.
\newblock In \emph{International Conference on Learning Representations}.

\bibitem[{Lu et~al.(2019)Lu, Batra, Parikh, and Lee}]{vilbert}
Lu, J.; Batra, D.; Parikh, D.; and Lee, S. 2019.
\newblock Vilbert: Pretraining task-agnostic visiolinguistic representations
  for vision-and-language tasks.
\newblock \emph{arXiv preprint arXiv:1908.02265}.

\bibitem[{Miller(1995)}]{wordnet}
Miller, G.~A. 1995.
\newblock WordNet: a lexical database for English.
\newblock \emph{Communications of the ACM}, 38(11): 39--41.

\bibitem[{Ordonez, Kulkarni, and Berg(2011)}]{sbu}
Ordonez, V.; Kulkarni, G.; and Berg, T.~L. 2011.
\newblock Im2Text: Describing Images Using 1 Million Captioned Photographs.
\newblock In \emph{Neural Information Processing Systems ({NIPS})}.

\bibitem[{Parekh et~al.(2020)Parekh, Baldridge, Cer, Waters, and Yang}]{CXC}
Parekh, Z.; Baldridge, J.; Cer, D.; Waters, A.; and Yang, Y. 2020.
\newblock Crisscrossed captions: Extended intramodal and intermodal semantic
  similarity judgments for MS-COCO.
\newblock \emph{arXiv preprint arXiv:2004.15020}.

\bibitem[{Plummer et~al.(2015)Plummer, Wang, Cervantes, Caicedo, Hockenmaier,
  and Lazebnik}]{flickr30k}
Plummer, B.~A.; Wang, L.; Cervantes, C.~M.; Caicedo, J.~C.; Hockenmaier, J.;
  and Lazebnik, S. 2015.
\newblock Flickr30k entities: Collecting region-to-phrase correspondences for
  richer image-to-sentence models.
\newblock In \emph{Proceedings of the IEEE international conference on computer
  vision}, 2641--2649.

\bibitem[{Qi et~al.(2020)Qi, Su, Song, Cui, Bharti, and Sacheti}]{imagebert}
Qi, D.; Su, L.; Song, J.; Cui, E.; Bharti, T.; and Sacheti, A. 2020.
\newblock {Imagebert: Cross-modal pre-training with large-scale weak-supervised
  image-text data}.
\newblock \emph{arXiv}, 1--12.

\bibitem[{Ren et~al.(2015)Ren, He, Girshick, and Sun}]{fasterRCNN:2015}
Ren, S.; He, K.; Girshick, R.; and Sun, J. 2015.
\newblock {Faster r-cnn: Towards real-time object detection with region
  proposal networks}.
\newblock In \emph{Advances in neural information processing systems}, 91--99.

\bibitem[{Selvaraju et~al.(2017)Selvaraju, Cogswell, Das, Vedantam, Parikh, and
  Batra}]{gradcam}
Selvaraju, R.~R.; Cogswell, M.; Das, A.; Vedantam, R.; Parikh, D.; and Batra,
  D. 2017.
\newblock Grad-cam: Visual explanations from deep networks via gradient-based
  localization.
\newblock In \emph{Proceedings of the IEEE international conference on computer
  vision}, 618--626.

\bibitem[{Sharma et~al.(2018)Sharma, Ding, Goodman, and Soricut}]{cc}
Sharma, P.; Ding, N.; Goodman, S.; and Soricut, R. 2018.
\newblock Conceptual captions: A cleaned, hypernymed, image alt-text dataset
  for automatic image captioning.
\newblock In \emph{Proceedings of the 56th Annual Meeting of the Association
  for Computational Linguistics (Volume 1: Long Papers)}, 2556--2565.

\bibitem[{Su et~al.(2019)Su, Zhu, Cao, Li, Lu, Wei, and Dai}]{vlbert}
Su, W.; Zhu, X.; Cao, Y.; Li, B.; Lu, L.; Wei, F.; and Dai, J. 2019.
\newblock VL-BERT: Pre-training of Generic Visual-Linguistic Representations.
\newblock In \emph{International Conference on Learning Representations}.

\bibitem[{Tan and Bansal(2019)}]{lxmert}
Tan, H.; and Bansal, M. 2019.
\newblock LXMERT: Learning Cross-Modality Encoder Representations from
  Transformers.
\newblock In \emph{Proceedings of the 2019 Conference on Empirical Methods in
  Natural Language Processing and the 9th International Joint Conference on
  Natural Language Processing}, 5100--5111.

\bibitem[{Vaswani et~al.(2017)Vaswani, Shazeer, Parmar, Uszkoreit, Jones,
  Gomez, Kaiser, and Polosukhin}]{transformer}
Vaswani, A.; Shazeer, N.; Parmar, N.; Uszkoreit, J.; Jones, L.; Gomez, A.~N.;
  Kaiser, {\L}.; and Polosukhin, I. 2017.
\newblock Attention is all you need.
\newblock \emph{Advances in neural information processing systems}, 30.

\bibitem[{Xie et~al.(2017)Xie, Girshick, Doll{\'a}r, Tu, and He}]{resnext}
Xie, S.; Girshick, R.; Doll{\'a}r, P.; Tu, Z.; and He, K. 2017.
\newblock Aggregated residual transformations for deep neural networks.
\newblock In \emph{Proceedings of the IEEE conference on computer vision and
  pattern recognition}, 1492--1500.

\bibitem[{Yang et~al.(2019)Yang, Dai, Yang, Carbonell, Salakhutdinov, and
  Le}]{infonce}
Yang, Z.; Dai, Z.; Yang, Y.; Carbonell, J.; Salakhutdinov, R.~R.; and Le, Q.~V.
  2019.
\newblock Xlnet: Generalized autoregressive pretraining for language
  understanding.
\newblock \emph{Advances in neural information processing systems}, 32.

\bibitem[{Yao et~al.(2021)Yao, Huang, Hou, Lu, Niu, Xu, and Xu}]{filip}
Yao, L.; Huang, R.; Hou, L.; Lu, G.; Niu, M.; Xu, H.; and Xu, C. 2021.
\newblock Fine-grained Interactive Language-Image Pre-Training.
\newblock In \emph{Proceedings of the International Conference on Learning
  Representations}.

\bibitem[{Yu et~al.(2022)Yu, Wang, Vasudevan, Yeung, Seyedhosseini, and
  Wu}]{coca}
Yu, J.; Wang, Z.; Vasudevan, V.; Yeung, L.; Seyedhosseini, M.; and Wu, Y. 2022.
\newblock Coca: Contrastive captioners are image-text foundation models.
\newblock \emph{arXiv preprint arXiv:2205.01917}.

\bibitem[{Yu et~al.(2018)Yu, Lin, Shen, Yang, Lu, Bansal, and
  Berg}]{yu2018mattnet}
Yu, L.; Lin, Z.; Shen, X.; Yang, J.; Lu, X.; Bansal, M.; and Berg, T.~L. 2018.
\newblock Mattnet: Modular attention network for referring expression
  comprehension.
\newblock In \emph{Proceedings of the IEEE Conference on Computer Vision and
  Pattern Recognition}, 1307--1315.

\bibitem[{Yu et~al.(2016)Yu, Poirson, Yang, Berg, and Berg}]{refcoco+}
Yu, L.; Poirson, P.; Yang, S.; Berg, A.~C.; and Berg, T.~L. 2016.
\newblock Modeling context in referring expressions.
\newblock In \emph{European Conference on Computer Vision}, 69--85. Springer.

\bibitem[{Zeng, Zhang, and Li(2021)}]{XITM}
Zeng, Y.; Zhang, X.; and Li, H. 2021.
\newblock Multi-Grained Vision Language Pre-Training: Aligning Texts with
  Visual Concepts.
\newblock \emph{arXiv preprint arXiv:2111.08276}.

\bibitem[{Zhang et~al.(2021)Zhang, Li, Hu, Yang, Zhang, Wang, Choi, and
  Gao}]{vinvl}
Zhang, P.; Li, X.; Hu, X.; Yang, J.; Zhang, L.; Wang, L.; Choi, Y.; and Gao, J.
  2021.
\newblock Vinvl: Revisiting visual representations in vision-language models.
\newblock In \emph{Proceedings of the IEEE/CVF Conference on Computer Vision
  and Pattern Recognition}, 5579--5588.

\bibitem[{Zhang et~al.(2020)Zhang, Zhao, Lin, He et~al.}]{ccl}
Zhang, Z.; Zhao, Z.; Lin, Z.; He, X.; et~al. 2020.
\newblock Counterfactual contrastive learning for weakly-supervised
  vision-language grounding.
\newblock \emph{Advances in Neural Information Processing Systems}, 33:
  18123--18134.

\end{thebibliography}

\end{document}